\documentclass[letterpaper,10pt,conference]{ieeeconf}
\IEEEoverridecommandlockouts
\usepackage{float}%
\usepackage{lipsum}
\usepackage{tablefootnote}
\usepackage{pifont}%

\usepackage[utf8x]{inputenc}

\usepackage{textcomp}
\usepackage{url}
\usepackage{hyperref}
\usepackage{amssymb}
\usepackage{graphicx}
\usepackage[cmex10]{amsmath}
\usepackage{array}
\usepackage{mdwtab,fixmath}
\usepackage{graphics}
\usepackage{epsfig}
\usepackage{times}
\usepackage{url}
\usepackage{lscape}
\usepackage{color}
\usepackage{epsfig}
\newtheorem{theorem}{Theorem}[section]
\newtheorem{remark}[theorem]{Remark}

\usepackage{cite}
\usepackage{balance}
\usepackage{dblfloatfix}
\usepackage{nicefrac}
\usepackage{xspace}

\allowdisplaybreaks

\usepackage{mdframed}
\usepackage{algorithm}
\usepackage{algorithmic}

\usepackage{tabularx}

\begin{document}

\title{\LARGE \bf Safe and Efficient Robot Action Planning in the Presence of Unconcerned Humans}

\author{Mohsen Amiri and Mehdi Hosseinzadeh, \IEEEmembership{Senior Member, IEEE}
\thanks{This research has been supported by National Science Foundation under award number DGE-2244082, and by WSU Voiland College of Engineering and Architecture through a start-up package to M. Hosseinzadeh.}
\thanks{The authors are with the Department of Mechanical and Materials Engineering, Washington State University, Pullman, WA 99164, USA. Email: mohsen.amiri@wsu.edu; mehdi.hosseinzadeh@wsu.edu.}
}

\maketitle

\begin{abstract}
This paper proposes a robot action planning scheme that provides an efficient and probabilistically safe plan for a robot interacting with an unconcerned human\textemdash someone who is either unaware of the robot's presence or unwilling to engage in ensuring safety. The proposed scheme is predictive, meaning that the robot is required to predict human actions over a finite future horizon; such predictions are often inaccurate in real-world scenarios. One possible approach to reduce the uncertainties is to provide the robot with the capability of reasoning about the human's awareness of potential dangers. This paper discusses that by using a binary variable, so-called danger awareness coefficient, it is possible to differentiate between concerned and unconcerned humans, and provides a learning algorithm to determine this coefficient by observing human actions. Moreover, this paper argues how humans rely on predictions of other agents' future actions (including those of robots in human-robot interaction) in their decision-making. It also shows that ignoring this aspect in predicting human's future actions can significantly degrade the efficiency of the interaction, causing agents to deviate from their optimal paths. The proposed robot action planning scheme is verified and validated via extensive simulation and experimental studies on a LoCoBot WidowX-250. 
\end{abstract}



\section{Introduction}
Safe and efficient interaction between humans and robots poses a significant challenge in Human\textendash Robot Interaction (HRI). Safety enforcement techniques often rely on assumptions that may not hold true in real-world scenarios. Consequently, robots might face incomplete or inaccurate information about their environment and other agents, particularly humans. For example, a human may be less attentive than usual when interacting with a self-driving car, assuming that the vehicle will handle all safety responsibilities. Addressing this uncertainty to maintain safety can compromise the robot's efficiency. Therefore, it is crucial for robots to effectively and promptly address environmental uncertainties to ensure both safety and efficiency in HRI.

To effectively interact with humans, robots need to predict human actions. Once a predictive human model is developed, the robot can leverage this model to create a safe and efficient plan by optimizing various high-level objectives \cite{wilcox2012,Ding2014}.

Regarding human's predictive model, some prior work (e.g., \cite{Huber2010}) assumes that robots possess complete knowledge of their environment; however, this assumption often falls short in real-world scenarios due to the inherent uncertainties in human behavior. To tackle this challenge, many researchers have focused on developing methods that enable robots to leverage historical human actions to predict future behaviors and states \cite{Kinugawa2017}. Various approaches, including Markov models, Gaussian mixture models, and Bayesian methods, have been employed in numerous studies to enhance the prediction of human actions; see, e.g., \cite{Amor2014}.



\begin{figure}[!t]
    \centering
    \includegraphics[width=0.48\linewidth]{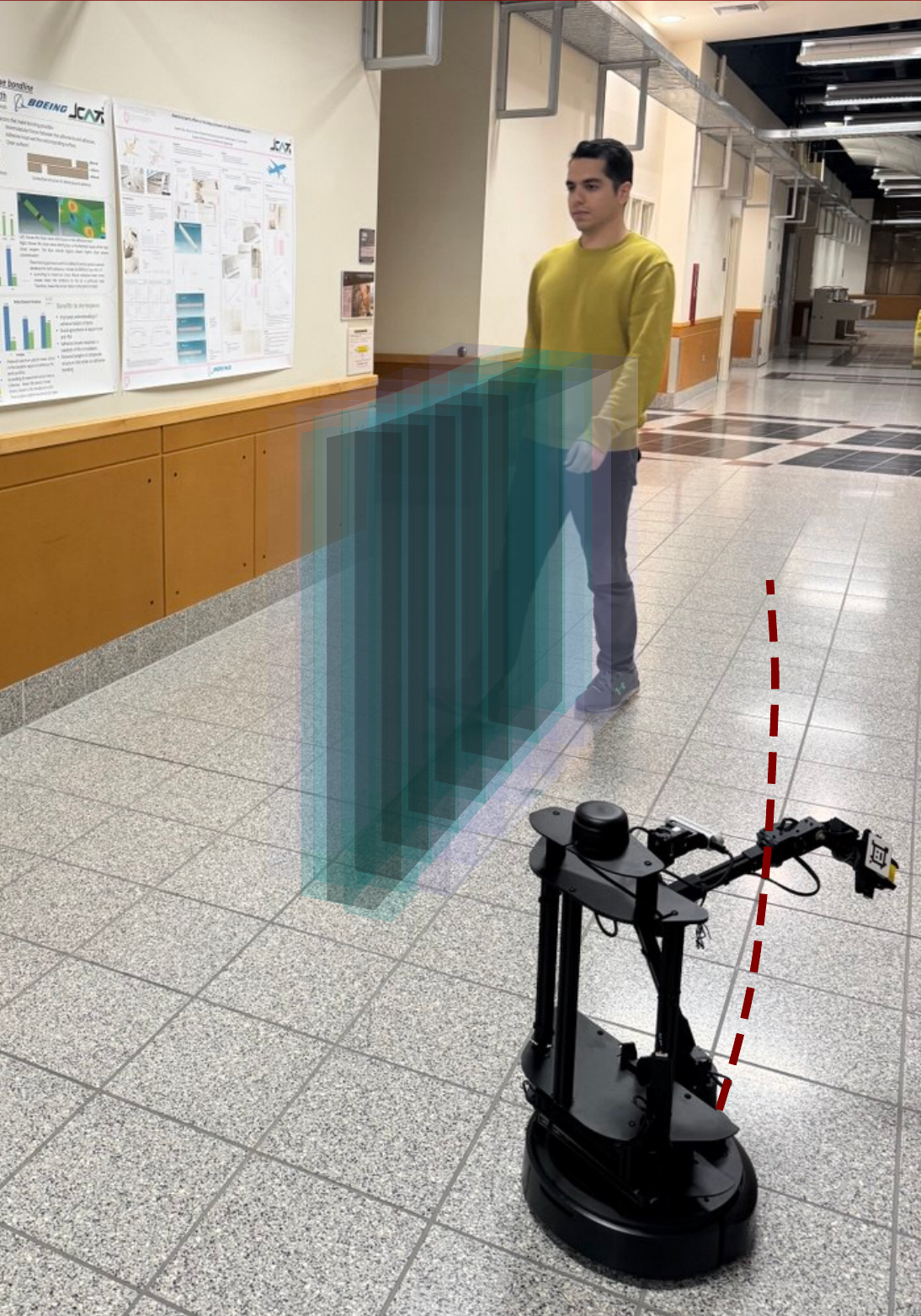}~~\includegraphics[width=0.48\linewidth]{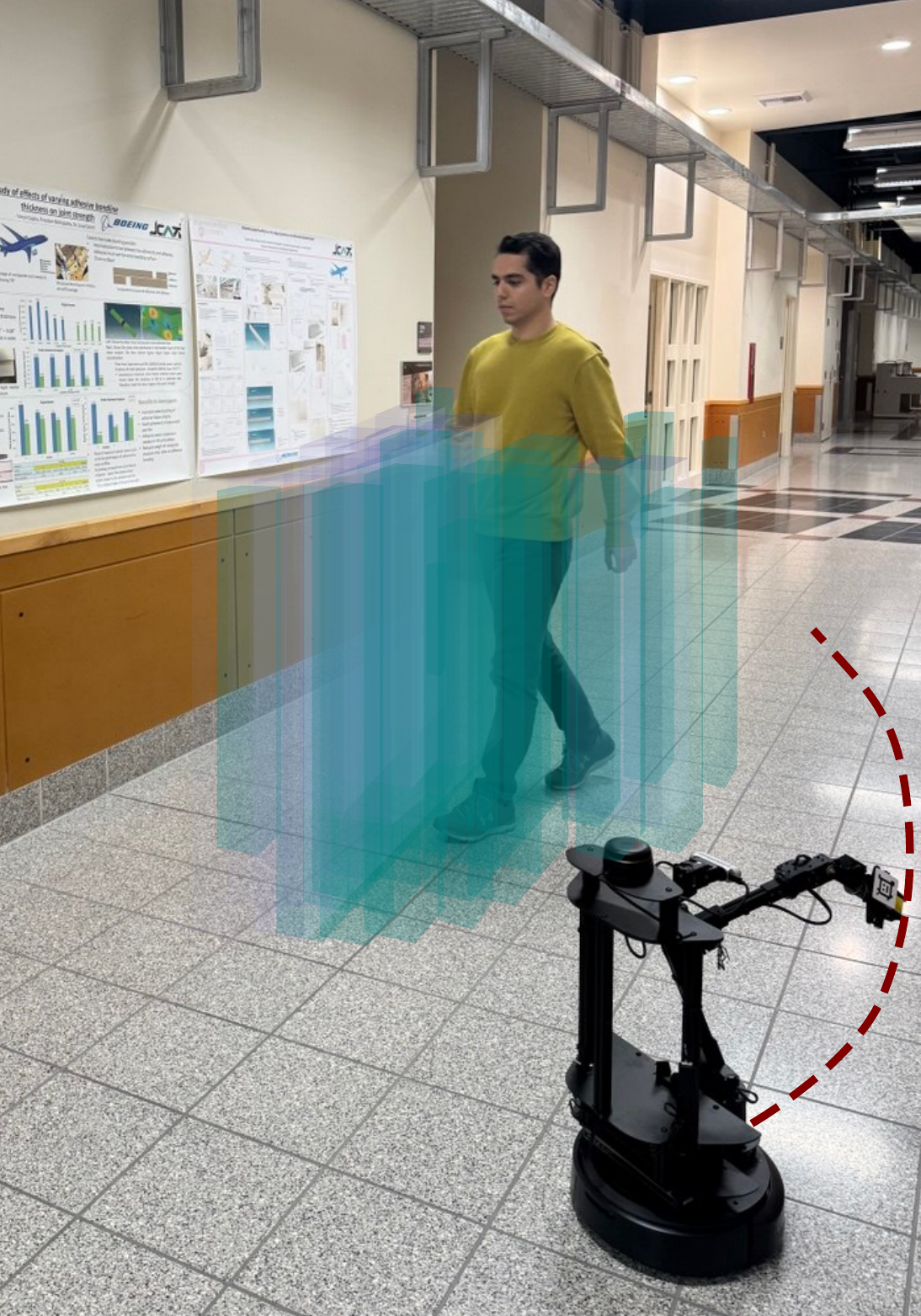}
    \caption{When developing action planning for robots that interact with humans, accurately predicting human actions is essential, as humans also predict the robot's actions in HRI environments. By taking human predictions into account, robots can improve their ability to predict human actions (left figure). Ignoring this aspect of human behavior can result in inaccurate predictions and reduced efficiency by causing the robot to deviate from its optimal path (right figure). A video can be viewed at \url{https://mhsnar.github.io/acc2025}}
    \label{fig:Figure1}
\end{figure}

Cognitive science literature offers an alternative approach for developing predictive models of human behavior. Research in this field (e.g., \cite{Baker2007,Neumann2007}) has shown that human actions can be effectively modeled using objective-driven optimization. Building on this framework, \cite{Fisac2018,Cheng2020} have developed models that predict future human actions by analyzing behavior during interactions. However, these models do not account for the human's danger awareness; specifically, whether the individual is aware of the robot's presence and any potential dangers, as well as their willingness to take steps to ensure safety. Note that by danger we refer to a situation in which the likelihood of a collision between the human and the robot is greater than a threshold value, if neither the robot nor the human change their behavior. Danger awareness among agents plays a crucial role in daily human interactions. For instance, a driver is more likely to pay attention to a cyclist in front of their car than one behind it, believing that the cyclist ahead may not be aware of the approaching vehicle. Similarly, a driver might slow down in an area where children are playing, while maintaining speed when adults are present, due to the understanding that children may inadvertently jump into the street. These examples illustrate how humans subconsciously consider each other's danger awareness in safety-critical situations. Therefore, incorporating the human's danger awareness is essential for any approach aimed at enabling robots to achieve human-level intelligence in ensuring safety.

With that, \cite{hosseinzadeh2023toward} quantified the notion of danger awareness as a binary variable and provided a game-theoretic interpretation to show that a binary variable is appropriate and sufficient to model the impact of human's danger awareness on their behavior. The paper also proposed a method for continuously learning a human's danger awareness by observing their actions. Additionally, it introduced a robot action planning scheme that integrates human danger awareness, providing a probabilistically safe and efficient planning strategy for the robot. Despite promising results, \cite{hosseinzadeh2023toward} evaluated the performance of the proposed scheme solely through simulation studies, which were confined to one-dimensional human and robot motions. Additionally, \cite{hosseinzadeh2023toward} suggested that human actions can be modeled using objective-driven optimization without the need for predictions. However, in reality, human actions are often driven by predictions \cite{huber2010assist}. For instance, when merging onto a highway, drivers evaluate the speed and distance of vehicles in adjacent lanes to predict whether they can safely join the traffic flow without causing collisions or requiring other drivers to adjust their speed. Similarly, drivers approaching a roundabout predict the movements of vehicles already in the roundabout and those waiting to enter, allowing them to choose an appropriate gap and merge safely.

The overall goal of this paper is to further advance the method described in \cite{hosseinzadeh2023toward} to expand its applicability and practicality, transitioning it from a lab setting to real-world applications. First, we develop a predictive model that predicts human actions and states by considering the predictive nature of human decision-making; see Fig. \ref{fig:Figure1}. Next, we develop a robot action planning scheme that generates a probabilistically safe and efficient plan by learning and incorporating human danger awareness. It is noteworthy that the proposed action planning scheme is not limited to one-dimensional motions, and it can accommodate realistic scenarios involving three-dimensional movements of both the robot and the human. Finally, we experimentally validate and evaluate the proposed scheme using a LoCoBot WidowX-250 across multiple real-world HRI scenarios.

\section{Problem Statement}\label{sec:ProblemStatement}
Consider a HRI scenario with one robot and one human, both moving towards distinct goal locations.

\medskip
\noindent\textbf{Human Model:} The human's movement is modeled as:
\begin{align}\label{eq:humanmodel}
x_{H}(t+1) &= f_{H}\left(x_{H}(t), u_{H}(t)\right)
\end{align}
where $ x_{H}(t) \in \mathbb{R}^{n_{H}} $ and $ u_{H}(t) \in \mathbb{R}^{m_{H}} $ are the human's state and action, respectively. We assume that the human uses a receding horizon framework to determine their action $u_H(t)$. Given $N_H\in\mathbb{Z}_{>0}$ as the length of the human's prediction horizon, the optimal sequence of human action $\mathbf{u_H}^\ast(t):=\left[\begin{matrix}\left(u_H^\ast(0 | t)\right)^{\top}& \cdots & \left(u_H^\ast(N_H-1 | t)\right)^{\top}\end{matrix}\right]^{\top} \in \mathbb{R}^{N_H\cdot m_H}$ over the prediction horizon can be determined by solving the following optimization problem:

\begin{subequations}\label{eq:HumanOP}
\begin{align}
 &\mathbf{u}_{H}^{\ast}(t)= \arg\,\min_{\mathbf{u}_{H}} \sum_{k=0}^{N_H-1}\left\|\hat{x}_{H}(k|t) - g_{H}\right\Vert_{\theta_{1}}^2\\ \nonumber
&~~~~~~~~+ \sum_{k=0}^{N_H-1}\left\|u_{H}(k|t)\right\Vert_{\theta_{2}}^{2}+\beta \sum_{k=0}^{N_H-1}\theta_{3}e^{-\theta_{4}\text{dist}(k|t)},\label{eq:costfunction}
\end{align}
subject to the following constraints:
\begin{align}
& \hat{x}_{H}(k+1|t)= f_{H}\left(\hat{x}_{H}(k|t), u_{H}(k|t)\right),~\hat{x}_{H}(0 | t)=x_{H}(t),\\
& \hat{x}_{H}(k | t) \in \mathcal{X}_H,~k \in\{0,1, \cdots ,N_H-1\},\\
& u_H(k | t) \in \mathcal{U}_H,~k \in\{0, \cdots, N_H-1\},
\end{align}
\end{subequations}
where $g_{H} \in \mathbb{R}^{n_{H}}$ is the human's goal state, $\hat{x}_{H}(k|t)$ represents state prediction at instant $k$ within the prediction horizon performed at time instant $t$, $\beta\in\{0,1\}$ is the \textit{danger awareness} coefficient (will be discussed in detail later), $\mathcal{X}_H\subset\mathbb{R}^{n_H}$ and $\mathcal{U}_H\subset\mathbb{R}^{m_H}$ are respectively the state and input constraints sets, $\text{dist}(k|t)$ is the Euclidean distance between robot and human at prediction instant $k$, and $\theta_{1}=\theta_{1}^{\top} \succeq 0$ ($\theta_{1} \in \mathbb{R}^{n_H \times n_H}$), $\theta_{2}=\theta_{2}^{\top} \succeq 0$ ($\theta_{2} \in \mathbb{R}^{m_H \times m_H}$)
, $\theta_{3}\in\mathbb{R}_{>0} $, and  $\theta_{4}\in\mathbb{R}_{\geq0}$ are weighting parameter. See \cite{hosseinzadeh2023toward} for a physical interpretation of $\beta$.

In \eqref{eq:costfunction}, the first and second terms are human's goal objective adopted from \cite{Bajcsy2019,hosseinzadeh2023toward}, and the last term is safety objective which is inspired from \cite{hosseinzadeh2023toward}. It should be noted that, as discussed in \cite{hosseinzadeh2023toward} and \cite{Fairfield2009}, human perception of distance from the robot is often noisy, which can be modeled by introducing normal noise with zero mean and constant covariance to the distance function $\text{dist}(k|t)$.

\medskip
\noindent\textbf{Robot Model:} The robot's movement is described by:
\begin{align}\label{eq:robotmodel}
x_{R}(t+1) &= f_{R}\left(x_{R}(t), u_{R}(t)\right)
\end{align}
where $ x_{R}(t) \in \mathbb{R}^{n_{R}} $ and $ u_{R}(t) \in \mathbb{R}^{m_{R}} $ are the robot's state and action at time instant $ t $, respectively.

\medskip
\noindent
\textbf{Goal:} Given the robot's goal state $g_R\in\mathbb{R}^{n_R}$, the primary objective of this paper is to develop a robot action planning scheme that determines the robot's action $u_R(t)$ to guide it towards its goal, while ensuring safety by avoiding collisions with humans.

\section{Proposed Robot Action Planning Scheme}\label{sec:ProposedSolution}

\medskip\noindent
\textbf{Human Action Prediction:} To predict human actions, the robot uses a mixture distribution model given by:
\begin{align}\label{eq:predictionFinal}
\mathbb{P}\big(u_H|x_H,x_R;\beta\big)=&(1-\omega_H)\cdot \mathbb{P}_d\big(u_H|x_H,x_R;\beta\big)\nonumber\\
&+\omega_H\cdot \mathbb{P}_r(u_H),
\end{align}
where $\mathbb{P}_d\big(u_H|x_H,x_R;\beta\big)$ represents the probability distribution of the human's deliberate actions, and $\mathbb{P}_r(u_H)$ accounts for random actions, which will be discussed in the following. The mixture weight $\omega_H\in[0,1]$ determines the influence of these components. Note that $\mathbb{P}\big(u_H|x_H,x_R;\beta\big)$ should be computed for every action $u_H\in\mathcal{U}_H$ and every danger awareness coefficient $\beta\in\{0,1\}$.

\medskip\noindent
\textit{Probability Distribution of Human's Deliberate Actions:} This probability distribution indicates whether the human chooses actions based on the goal and safety objective functions detailed in\eqref{eq:HumanOP}. The probability distribution of the human's deliberate actions at any time instant $t$ can be computed as:
\begin{align}\label{eq:prediction}
\mathbb{P}_d\big(u_H|x_H,x_R;\beta\big)=& \left\{ 
\begin{array}{ll}
1, & ~~~\text{if} ~ u_H=u_H^{\ast}(0|t) \\
0, &~~~ \text{otherwise}
\end{array}
\right.,
\end{align}
where $u_H^{\ast}(0|t)$ is as in \eqref{eq:HumanOP}. Under the assumptions described in Remark \ref{remark1}, the robot can compute the human's optimal action $u_H^{\ast}(0|t)$ at any given time instant.

\medskip\noindent
\textit{Probability Distribution of Human's Random Actions:} This probability distribution reflects whether the human chooses actions randomly, completely disregarding the above-mentioned goal and safety objective functions. At any time instant $t$, the robot uses a uniform distribution to represent the probability of the human's random actions, as:
\begin{align}\label{eq:PrHuman}
\mathbb{P}_r(u_H)=\frac{1}{\vert\mathcal{U}_H\vert},~\forall u_H\in\mathcal{U}_H,
\end{align}
where $\vert\mathcal{U}_H\vert$ is the cardinality of the set $\mathcal{U}_H$.

\begin{remark}
The choice of the discrete action space $\mathcal{U}_H$ is crucial for the effectiveness of robot action planning schemes. The set $\mathcal{U}_H$ should be a reasonable approximation of the range of human actions in the given interaction. While there are methods for defining $\mathcal{U}_H$ for specific applications (e.g., "action point" models \cite{Michaels1963, Brackston1999} used in \cite{Li2019} to define a human driver's action set), $\mathcal{U}_H$ is often determined empirically in most prior work. In our experiments, we intuitively design the human action space to encompass all possible actions in a realistic scenario.  
\end{remark}

\medskip\noindent
\textbf{Computation of the Collision Probability:} At any time instant $t$, the robot can predict the human's future states based on their current state and its predictions of the human's future actions, and calculate the collision probability at future time instants. This will be discussed further below.

Let the $x_H$-space be divided into $N_c$ grid cells. The probability of a collision at prediction instant $k$ can then be computed as the probability that $x_R(k)$ and $x_H(k)$ are within the same grid cell, assuming no collisions occur prior to $k$. Thus, to compute the collision probability, the robot needs to compute the probability distribution of the human’s states over the prediction horizon $[t,t+N_R]$, where $N_R$ is the length of the prediction horizon.

\begin{remark}\label{remark:NR}
In this paper, we assume that $N_R=N_H$, where $N_H$ is the length of the prediction horizon used by the human for decision-making. Future work will explore other scenarios and assess their impact on system performance.
\end{remark}

The robot uses the following recursive process to compute the probability distribution of human's states over the prediction horizon $[t,t+N_R]$:
\begin{align}\label{eq:stateprediction}
&\mathbb{P}(x_H(k+1))\propto\sum\limits_{x_H(k),u_H(k),\beta}\mathbb{P}\left(x_H(k+1)|x_H(k),u_H(k)\right)\nonumber\\
&~~~~~~~~~~~~~~~~\cdot\mathbb{P}\left(u_H(k)|x_H(k),x_R(k);\beta\right)\cdot\mathbb{P}_t(\beta),
\end{align}
where $k=t,\cdots,t+N_R-1$, $\mathbb{P}\big(u_H(k)|x_H(k),x_R(k);\beta\big)$ is as in \eqref{eq:predictionFinal}, $\mathbb{P}_t(\beta)$ is the robot's belief about the human's danger awareness coefficient (that will be discussed later), and $\mathbb{P}\big(x_H(k+1)|x_H(k),u_H(k)\big)$ is equal to the previous probability of human's states in $x_H(k)$ if $x_H(k+1)$, $x_H(k)$, and $u_H(k)$ satisfy \eqref{eq:humanmodel}, and is equal to zero otherwise.

\medskip\noindent
\textbf{Robot's Belief about the Human's Danger Awareness Coefficient:} Let $\mathbb{P}_t(\beta=1)$ represent the robot's belief at time instant $t$ about the probability that the human is concerned, where $\mathbb{P}_0(\beta=1)$ denotes the robot's prior belief. The robot's belief about the probability that the human is unconcerned can then be computed as $\mathbb{P}_t(\beta=0)=1-\mathbb{P}_t(\beta=1)$. 

At any time instant $t$, by observing the human's state $x_H(t)$ and action $u_H(t)$, the robot updates its belief using the following Bayesian update rule:
\begin{align}\label{eq:updatebelief}
\mathbb{P}_{t+1}(\beta=1)=\frac{\mathbb{P}(u_H(t)|x_H(t),x_R(t);\beta=1)\mathbb{P}_t(\beta=1)}{\sum\limits_{\beta}\mathbb{P}(u_H(t)|x_H(t),x_R(t);\beta)\mathbb{P}_t(\beta)},
\end{align}
where $\mathbb{P}(u_H(t)|x_H(t);\beta=1)$ is equivalent to $\mathbb{P}(u_H=u_H(t)|x_H(t);\beta=1)$ which is computed in \eqref{eq:predictionFinal}.

\medskip
\noindent\textbf{Robot Action Planning:}
The robot uses a receding horizon control strategy to find the optimal control sequence $\mathbf{u}_R^\ast(t):=\left[\begin{matrix}\left(u_R^\ast(0 | t)\right)^{\top}& \cdots & \left(u_R^\ast(N_R-1 | t)\right)^{\top}\end{matrix}\right]^{\top} \in \mathbb{R}^{N_R\cdot m_R}$  by solving the following optimization problem:
\begin{subequations}\label{eq:RobotOP}
\begin{align}
\mathbf{u}_{R}^{\ast}\left(t\right) &= \arg\,\min_{\mathbf{u}_{R}} \sum_{k=0}^{N_R-1}\left\|\hat{x}_{R}(k|t) - g_{R}\right\Vert_{\theta_{5}}^{2}\\ \nonumber
&+\sum_{k=0}^{N_R-1}\left\|u_{R}(k|t)\right\Vert_{\theta_{6}}^{2},\label{eq:costfunction}
\end{align}
subject to the following constraints:
\begin{align}
& \hat{x}_{R}(k+1|t)= f_{R}\left(\hat{x}_{R}(k|t), u_{R}(k|t)\right),~\hat{x}_R(0 | t)=x_R(t),\\
& \hat{x}_{R}(k | t) \in \mathcal{X}_R,~k \in\{0,1, \cdots ,N_R-1\},\\
& u_R(k | t) \in \mathcal{U}_R,~k \in\{0,1, \cdots, N_R-1\},\\
& \mathbb{P}_{\text{Coll}}(k|t)\leq P_{th},~k \in\{0,1 \cdots, N_R-1\},\label{eq:P_Coll}
\end{align}
\end{subequations}
where $N_R\in\mathbb{Z}_{>0}$ is the length of the prediction horizon (see Remark \ref{remark:NR}), $\hat{x}_{R}(k|t)$ represents state prediction at instant $k$ within the prediction horizon performed at time instant $t$, $\mathcal{X}_R\subset\mathbb{R}^{n_R}$ and $\mathcal{U}_R\subset\mathbb{R}^{m_R}$ are respectively the state and input constraints sets, $ \mathbb{P}_{\text{Coll}}(k) \in [0,1] $ is the probability of a collision between the human and robot at prediction instant $ k $ (this can  be determined based on the probability distribution of human's future states as discussed above), $\theta_{5}=\theta_{5}^{\top} \succeq 0$ ($\theta_{5} \in \mathbb{R}^{n_R \times n_R}$)
, $\theta_{6}=\theta_{6}^{\top} \succeq 0$ ($\theta_{6} \in \mathbb{R}^{m_R \times m_R}$), and $ P_{\mathrm{th}} \in [0,1] $ is a threshold value.



\section{Investigating the Impact of Human's Predictive Behavior}\label{sec:simulation}

This section investigates the impact of considering predictions in human decision-making on efficiency of HRI. Note that unlike the unrealistic assumption in \cite{hosseinzadeh2023toward} that the robot is limited to vertical motion, we model both human and robot movements in the X-Y plane. This assumption aligns perfectly with real-world scenarios where a ground robot interacts with a human in various environments, such as streets or orchards. 


We adopted the model described in \cite{Zhang2023,Luca1998} to describe the dynamics of the LoCoBot WidowX-250: 
\begin{equation}\label{eq:ModelRobot}
x_R(t+1) = \begin{bmatrix} 1 & 0 \\ 0 & 1 \end{bmatrix}x_R(t) +\begin{bmatrix} \Delta T & 0 \\ 0 & \Delta T \end{bmatrix} u_R(t),
\end{equation}
where \(x_R(t) \in \mathbb{R}^2\) represents the robot's position at time instant \(t\), \(u_R(t) \in \mathbb{R}^2\) denotes its velocity, and \(\Delta T\) is the sampling period. We assume that $u_R$ can take any value in the range $[-2,2]\times[-2,2]$.

Similar to \cite{Fisac2018,hosseinzadeh2023toward}, the human's dynamical model is:
\begin{equation}\label{eq:ModelHuman}
x_H(t+1) = \begin{bmatrix} 1 & 0 \\ 0 & 1 \end{bmatrix} x_H(t) +\begin{bmatrix} \Delta T & 0 \\ 0 & \Delta T \end{bmatrix} u_H(t) ,
\end{equation}
where \(x_H(t) \in \mathbb{R}^2\) represents the human's position at time instant \(t\), and \(u_H(t) \in \mathbb{R}^2\) denotes their velocity. The set of control inputs for the human is defined as \(\mathcal{U}_H = \{-2v_H, -v_H, 0, v_H, 2v_H\}\times\{-2v_H, -v_H, 0, v_H, 2v_H\}\), where \(v_H=0.5\) [m/s] is the human's nominal walking speed. This set allows the pedestrian to stop, walk, or run in any direction on the X-Y plane, and captures almost all human's actions in real-world scenarios.

In this section, we assume that $\beta=1$ (i.e., the human is concerned), \(g_R = [0,10]^T\), \(g_H = [5,0]^T\), \(v_H = 0.5\), \(\omega_H = 0.5\), \(P_{\mathrm{th}} = 0.05\), \(\theta_1 =\text{diag}\{1,1\}\), \(\theta_2 = \text{diag}\{1,1\}\), \(\theta_3 = 2.5\), \(\theta_4 = 8 \times 10^{-3}\), \(\theta_5 = \text{diag}\{100,100\}\), \(\theta_6 = \text{diag}\{0.06,0.06\}\),  and \(\mathbb{P}_0(\beta = 1) = 0.5\).  We also assume that the initial conditions are $x_R(0)=[0~-10]^\top$ and $x_H(0)=[-5~0]^\top$. To implement the constraint \eqref{eq:P_Coll}, we require the robot to maintain a distance of 0.5 meters from all grid cells in which $P_{\text{Coll}}$ is greater than or equal to $P_{th}$. The \texttt{SciPy} toolbox is employed to solve the resulting optimization problems. 


\subsection{Impact of Incorporating Predictions in Human's Decision-Making}\label{sec:ComparisonPx}
\begin{figure}[!t]
    \centering \includegraphics[width=\columnwidth]{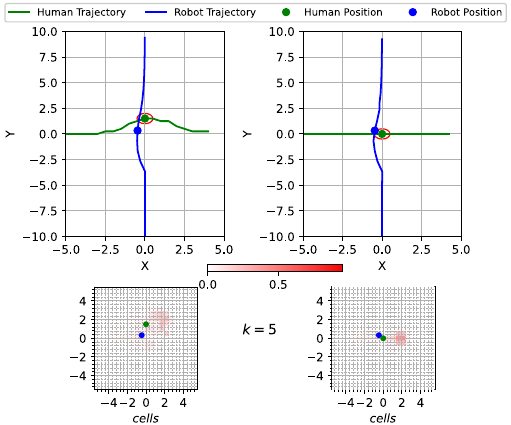}
    \caption{Impact of incorporating predictions in human's decision-making. The left column presents results without predictions and the right column shows results with predictions. The bottom figures present the probability distribution of human's future states at each prediction instant using the red grid. Not that in both top figures, the robot moves vertically from bottom to top, while the human moves horizontally from left to right.}
    \label{fig:Comparison}
\end{figure}

The probability distributions of human states with ($N_H=5$) and without ($N_H=0$) predictions in human's decision-making is shown in Fig. \ref{fig:Comparison}. As seen in this figure, when humans incorporate predictions into their decision-making, the trajectories of both the robot and the human become smoother and shorter. In other words, removing the predictive aspect of human decision-making creates challenges and difficulties for both the human and robot. This observation is reasonable, as the human predicts potential dangers and actively works to mitigate them. More precisely, when the human does not take into account future predictions, the probability distribution over their states in the future, which is computed via  \eqref{eq:stateprediction}, become wide, as shown in Fig. \ref{fig:Comparison}; this results in the robot adopting a more conservative behavior. Note that although we reported the results for $N_H=5$ as it is the optimal prediction horizon length for this scenario (see the following subsection), the observation is true for other values of $N_H$.

The discussion above highlighted that incorporating predictions into human decision-making not only enhances the realism of experiments and aligns them more closely with real-world scenarios, but also leads to improved performance and greater efficiency in HRI.

\subsection{Impact of the Length of the Human's Prediction Horizon on Interaction Efficiency} \label{sec:ComparisonPerformance}
To examine the impact of $N_H$ on the HRI efficiency, we consider the following  performance indices: $PI_R:= \sum_{t}\left\Vert x_R(t)- g_R \right\Vert^2$ which is the deviations of the robot from its minimum distance path to its goal;  $PI_H:= \sum_{t}\left\Vert x_H(t) - g_H \right\Vert^2$ which is the deviations of the human from their minimum distance path to their goal; and $PI_T:=PI_R+PI_H$.

\begin{figure}[!t]
    \centering
    \includegraphics[width=.7\linewidth]{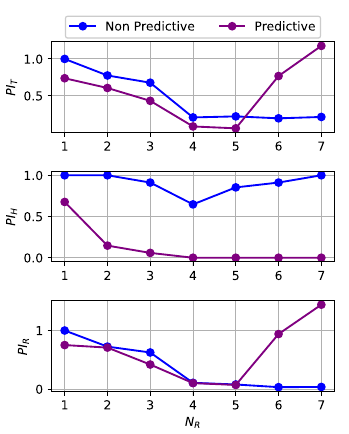}
    \caption{The impact of $N_R$ on  performance indices, both with and without consideration of the predictive aspect of human behavior.}
    \label{fig:PI}
\end{figure}

Fig. \ref{fig:PI} presents the obtained performance indices for different values of $N_R$, where in the non-predictive cases $N_H$ is set to zero and in predictive cases $N_H$ is equal to $N_R$. In this figure, the index corresponding to $N_R=1$ for the non-predictive cases is used as the basis for normalization. From Fig. \ref{fig:PI}, it can be seen that when the robot's planning does not take into account the predictive aspect of human decision-making, increasing the prediction horizon length (i.e., increasing $N_R$) reduces the robot's performance index $PI_R$ and the overall index $PI_T$ significantly. However, the impact on the human's performance index $PI_H$ is minimal. Furthermore, if $N_R$ continues to increase, the performance index $PI_T$ remains constant. This observation reveals that neglecting the full spectrum of human decision-making cannot be compensated for by simply enhancing the robot's predictive capabilities through a larger $N_R$.

When the robot's planning accounts for the predictive aspect of human decision-making, Fig. \ref{fig:PI} shows that increasing $N_H$  leads to a reduction in $PI_R$ and $PI_T$. This reduction occurs because the robot is able to incorporate more information about future conditions. However, as we look further into the future (i.e., as $N_H$ continues to increase), $PI_R$ and $PI_T$ increase due to reduced prediction accuracy resulting from the human's random actions described in \eqref{eq:PrHuman}. For what regards $PI_H$, increasing $N_R$ (i.e., increasing the length of the robot's prediction horizon) results in a decrease in $PI_H$, indicating that a larger $N_R$ can help reduce deviations in the human's trajectory from the optimal path. Note that despite $PI_R$ and $PI_T$, further increasing the $N_R$ does not lead to an increase in $PI_H$. However, it is obvious that with increasing prediction lengths, both $PI_T$, and $PI_R$ increase, as the complexity and uncertainty of the problem grows due to the unnecessarily large prediction horizon, primarily originating from the robot's contribution.

\section{Experimental Results}\label{sec:experiment}
This section aims at experimentally validating the proposed scheme by investigating its performance in providing a safe path to the LoCoBot WidowX-250 interacting with a human. We assume a scenario in which the human and robot walk toward their goal positions, while moving toward each other. We use models given in \eqref{eq:ModelRobot} and \eqref{eq:ModelHuman} to describe the dynamical behavior of the human and robot, respectively. We assume that $g_R=[0,~-3]^\top$, $g_H=[-2,~0]^\top$, $x_R(0)=[0~3]^\top$, $P_{\mathrm{th}} = 0.05$, $x_H(0)=[2~0]^\top$, $N_R=2$, and $\mathcal{U}_R=[-0.1,0.1]\times[-0.1,0.1]$; other parameters are as in Section \ref{sec:simulation}. 

\begin{figure}
    \centering
    \includegraphics[width=\linewidth]{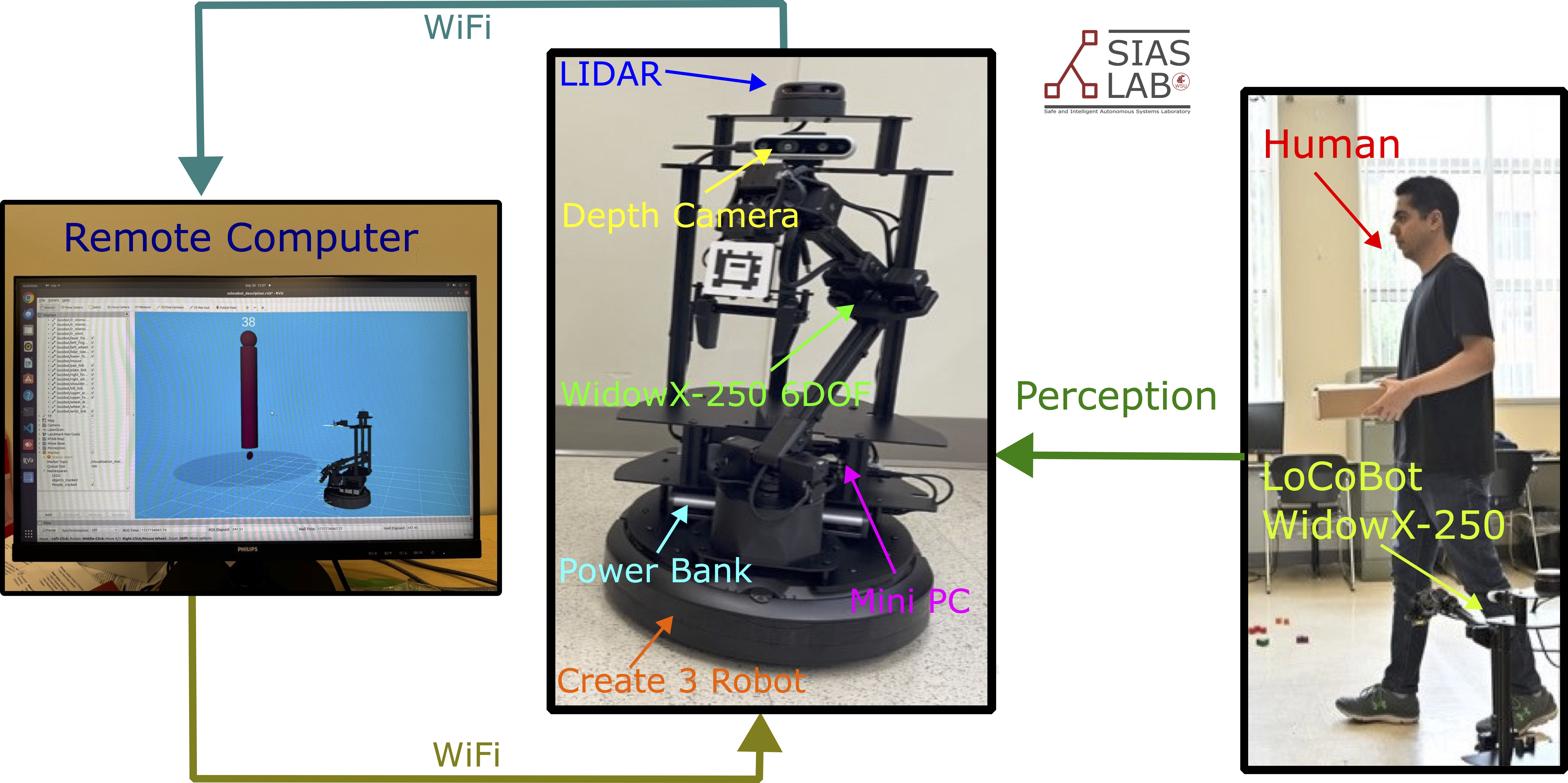}
    \caption{Overview of the experimental setup utilized to experimentally validate the proposed robot action planning scheme.}
    \label{fig:network}
\end{figure}

Our experimental setup (see Fig. \ref{fig:network}) includes a remote control system that initializes and monitors the whole process and LoCoBot WidowX-250. The robot's computing unit is a $8^{\text{th}}$ Generation $\text{Intel}^{\text{\textregistered}}$ $\text{Dual-Core}^{\text{\texttrademark}}$ i3-processor coupled with 8GB DDR4 RAM, which analyzes and interprets data while performing the computations of the proposed scheme. Communication between the robot's computing unit and the remote computer is established through WiFi. We use a RPLidar A2M8 to detect the human and measure their position and velocity at any time, providing 2D scans within a range of 10 meters. We implemented the leg tracker detailed in \cite{leigh2015person} to interpret the LiDAR data and determine the human's position and velocity. The entire system operates at the frequency of 2 Hz, while the LiDAR runs at the frequency of 10 Hz. To solve the optimization problem \eqref{eq:RobotOP}, we used the warm starting method described in \cite{Hosseinzadeh2023RobustTermination,AMIRI2024330,amiri2025practical,SSMPC}.



\begin{figure}[t]
    \centering
    \includegraphics[width=1\linewidth]{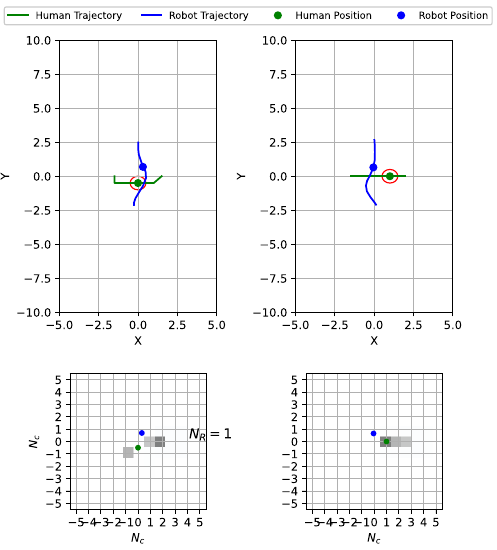}
    \caption{The left column presents experimental results for concerned human  and the right column
shows results for unconcerned human, while both are using predictions in human's decision-making. The second row of figures present the probability distributing of human’s future states for future time instants using the red grid. Not that in both top figures, the robot moves vertically from bottom to top, while the human moves from top to bottom.}
    \label{fig:EXp}
\end{figure}

Experimental results are presented in Fig. \ref{fig:EXp}, where the left and right figures show results for a concerned and unconcerned human, respectively. From this figure, we see that the proposed scheme can effectively distinguish between concerned and unconcerned humans and act accordingly. Fig. \ref{fig:EXp} shows that when the human is unconcerned, the robot must deviate further from its minimum distance path to ensure safety, which reduces its efficiency. In particular, the normalized performance index $PI_R$ for the unconcerned human is 1.62, while it is 1 for the concerned human; that is, unconcerned human results in an extra 62\% deviation in robot's path compared to a concerned human. Regarding the probability distribution over the human's future states, as the robot effectively learns the human's danger awareness coefficient, the distribution becomes narrower for both concerned and unconcerned humans, indicating that uncertainty about the human's future states is minimized.


\section{Conclusion}\label{sec:conclusion}
This paper proposes a robot action planning scheme capable of providing a safe and efficient plan for robots interacting with unconcerned humans who are either unaware of potential dangers (e.g., children and individuals with visual impairments) or unwilling to cooperate with the robot in ensuring safety. To achieve this, first, it was shown that concerned and unconcerned humans can be discriminated using a binary variable, and a Bayesian learning method was developed to determine this variable by observing human actions. It was then argued that humans consider predictions of others' actions, and it was shown that ignoring this aspect of human decision-making can significantly degrade the efficiency of HRI. Finally, a predictive robot action planning scheme was developed by incorporating the aforementioned discussions and concepts to predict human actions over a finite future horizon. Extensive simulation and experimental results showed that the proposed scheme can effectively reason over the uncertainties in predicting human actions, and provide a probabilistically safe plan for the robot. 

At the current stage, the main obstacles to the widespread adoption of the proposed methodology is its high computational complexity, which becomes significantly more challenging as the length of the prediction horizon increases. To the best of our knowledge, this issue is not unique to our planner, but to all planning schemes that take into account human's behavioral factors, e.g., \cite{Keil2020}. Future work will focus on developing robot action planning schemes with a lower computational footprint. We are also interested in extending this methodology to multi-robot, multi-human settings. Additionally, we believe that the proposed methodology can be integrated with other planning schemes that consider human behavioral factors.

\section*{Acknowledgment}
The authors would like to thank Dr. John P. Swensen for his insightful advice on the experimental setup.



\bibliographystyle{IEEEtran.bst} 
\bibliography{ref}  

\end{document}